\documentclass[letterpaper]{article} 
\usepackage{aaai25}  
\usepackage{times}  
\usepackage{helvet}  
\usepackage{courier}  
\usepackage[hyphens]{url}  
\usepackage{graphicx} 
\usepackage{natbib}  
\usepackage{caption} 
\usepackage{algorithm}
\usepackage{algorithmic}
\usepackage{amsmath,amsfonts}
\usepackage{multirow}
\usepackage{booktabs}
\usepackage{tabularx}
\usepackage{makecell}
\usepackage{newfloat}
\usepackage{listings}
\DeclareCaptionStyle{ruled}{labelfont=normalfont,labelsep=colon,strut=off} 
\floatstyle{ruled}
\newfloat{listing}{tb}{lst}{}
\floatname{listing}{Listing}
\title{Noise-Resilient Symbolic Regression with Dynamic Gating Reinforcement Learning}
\author{
    Chenglu Sun\textsuperscript{\rm 1},
    Shuo Shen\textsuperscript{\rm 1},
    Wenzhi Tao\textsuperscript{\rm 1},
    Deyi Xue\textsuperscript{\rm 1},
    Zixia Zhou\textsuperscript{\rm 2}\thanks{Corresponding Author: Zixia Zhou}
}
\affiliations{
    \textsuperscript{\rm 1}Cooperation Product Department, Interactive Entertainment Group, Tencent\\
    \textsuperscript{\rm 2}Stanford University\\

    \{clivesun, svenshen, simontao, davexue\}@tencent.com, zixia@stanford.edu
}

\usepackage{bibentry}

\begin{document}

\maketitle

\begin{abstract}
Symbolic regression (SR) has emerged as a pivotal technique for uncovering the intrinsic information within data and enhancing the interpretability of AI models. However, current state-of-the-art (sota) SR methods struggle to perform correct recovery of symbolic expressions from high-noise data. To address this issue, we introduce a novel noise-resilient SR (NRSR) method capable of recovering expressions from high-noise data. Our method leverages a novel reinforcement learning (RL) approach in conjunction with a designed noise-resilient gating module (NGM) to learn symbolic selection policies. The gating module can dynamically filter the meaningless information from high-noise data, thereby demonstrating a high noise-resilient capability for the SR process. And we also design a mixed path entropy (MPE) bonus term in the RL process to increase the exploration capabilities of the policy. Experimental results demonstrate that our method significantly outperforms several popular baselines on benchmarks with high-noise data. Furthermore, our method also can achieve sota performance on benchmarks with clean data, showcasing its robustness and efficacy in SR tasks.
\end{abstract}

\section{Introduction}
\label{Section:1}

With the rise of electronic information technology, we have easy access to abundant data for acquisition, processing, and analysis. Extracting meaningful relationships from data is crucial for AI design, scientific discovery, and identifying core factors, et al. Deep learning (DL) has emerged as a powerful tool for data mining \cite{shu2023knowledge,sorscher2022beyond}, enabling neural networks to tackle a wide array of scientific tasks, such as regression and classification problems \cite{muthukumar2021classification}. For instance, given a dataset $(X, y)$, where each point $X_i \in \mathbb{R}^n$ and $y_i \in \mathbb{R}$, DL methods can train a network to approximate $y_i\approx F(X_i)$, with $F$ representing the learned network. However, as a black-box system \cite{buhrmester2021analysis}, the relationships established by the network between data and targets are often opaque, making fine-grained control of the system and understand the information contained in the data challenging. Therefore, improving model interpretability and controllability has gradually become a key research direction in the current AI field.

Symbolic regression (SR) stands out as a promising technology, aiming to discover the correct symbolic expression for an unknown function $f$ that best fits the dataset with $y_i = f(X_i)$. SR has the potential to explain black-box systems with simple, accurate, intelligible expressions, enhancing the development of data-driven AI systems \cite{udrescu2020ai,oliveira2018analysing,jiang2023symbolic, udrescu2020ai2,kim2020integration,petersen2021deep,kamienny2022end,kamienny2023deep}. Despite its promise, SR is difficult due to the exponentially large combinatorial space of symbolic expressions. Traditional SR methods are usually based on genetic programming (GP) \cite{schmidt2009distilling,searson2010gptips,de2014kaizen}. While GP-based SR method achieve good performance, they are slow, and struggle to scale to larger problems and are highly sensitive to hyperparameter settings \cite{mundhenk2021seeding}. Recent advances have seen a pivot towards DL approaches for SR, leveraging networks to represent and learn the semantics of symbols \cite{kamienny2022end,kim2020integration,Ascoli2022deep}. However, DL-based methods often encounter difficulties in achieving satisfactory regression performance or need specific revisions on the network and search space. Some DL-based methods pre-train an encoder-decoder network to learn the expression representations within a given dataset \cite{valipour2021symbolicgpt, holt2023deep}. These methods sample function $f$ using the pre-trained network, thereby achieving low complexity at inference process. But the pre-trained model may lead to sub-optimal solutions and fail to discover highly complex equations \cite{holt2023deep}.

Reinforcement learning (RL) \cite{wang2022deep,nguyen2020deep,crochepierre2022interactive,landajuela2021discovering} applied to SR combines the benefits of GP with environmental feedback and the representation power of neural networks. This combination has contributed to a growing trend in research adopting RL for SR tasks \cite{mundhenk2021symbolic, petersen2021deep, landajuela2022unified,zhang2021rl}. However, a significant portion of real-world data is characterized by high-noisy\footnote{In this paper, ``noises'' refers to irrelevant variables within a dataset.}, often containing abundant noisy information or irrelevant data. Performing SR on such high-noise data is exceedingly complex due to the dramatic increase in the search space caused by the noisy input variables \cite{reinbold2021robust}. Yet, most benchmarks used in SR research are typically clean, posing a significant challenge for the application of these algorithms in real-world scenarios. To address this issue, we introduce a novel end-to-end Noise-Resilient SR method, termed as NRSR, which can learn expressions from high-noise data via RL with a Noise-Resilient Gating Module (NGM). The NGM, designed to dynamically filter out noisy input variables during the RL training process, significantly boosts the efficiency of exploration. Additionally, we propose a Mixed Path Entropy (MPE) regularizer to bolster the exploration capabilities for searching expressions and prevent overfitting of the RL model. Hence, NRSR not only accurately recovers expressions from high-noise data but also enhances the precision of RL-based SR approaches. To evaluate the effectiveness of NRSR, we employed a suite of benchmarks containing twelve representative expressions and selected five popular SR approaches as baselines for a comprehensive test. The results demonstrate that NRSR significantly outperforms on benchmarks with high-noise data and can surpass all baselines on benchmarks with clean data. The main contributions of this study are threefold: (1) We introduce NRSR, a novel SR method that exhibits state-of-the-art (sota) performance on both high-noise data and clean data. (2) We design a dynamic NGM that effectively filters noisy variables, and the proposed MPE enhances the model's exploration ability in expression generation. (3) Through experiments, we analyze the performance of the NGM and the MPE regularizer. These components can be decoupled from our method, allowing for integration with other scenarios.

\section{Related Works}
\label{Section:2}

\subsubsection{Reinforcement learning for symbolic regression} Several recent approaches leverage RL-based method for SR, and represents decent performance. Petersen et al. \cite{petersen2021deep} introduced an SR framework utilizing RL, where a RNN generates mathematical expressions optimized through a risk-seeking policy gradient (PG) algorithm, outperforming established baselines and commercial software on twelve benchmarks. Crochepierre et al. \cite{crochepierre2022interactive} proposed an interactive web-based platform that enhances grammar-guided SR by incorporating user preferences through a RL framework. Zhang et al. \cite{zhang2021rl} presented an SR method that combines genetic algorithms and RL to solve SR problems, addressing the challenge of balancing exploration and exploitation. The experimental results, based on ten benchmarks, demonstrate that this hybrid approach achieves competitive performance. And other existing RL-based SR methods also have shown promising results \cite{landajuela2021discovering,mundhenk2021symbolic,landajuela2022unified}. RL possesses robust decision-making capabilities and an exceptional aptitude for target fitting, making it an ideal foundational path for SR. Despite the advancements in RL-based SR methods, their performance on high-noise data has not been satisfactory. This area, therefore, warrants further exploration to improve the robustness and applicability of these techniques.

\subsubsection{L0 Regularization} L1 and L2 regularization are commonly used regularization methods in neural networks \cite{ma2019transformed, cortes2012l2}, which can enhance the generalization of the model and prevent overfitting \cite{ying2019overview}. L1 and L2 regularization respectively limit the absolute value and squared magnitude of the weights in the network. L0 regularization is to encourage the network to have a small number of non-zero parameters \cite{louizos2018learning,wei2022neural}, which can lead to sparser models. However, L0 regularization is non-differentiable and thus harder to implement in practice. Louizos et al. \cite{louizos2018learning} introduced an L0-norm regularization technique for neural networks that employs stochastic units to induce sparsity, enabling differentiable pruning during training, which enhances generalization. Wei et al. \cite{wei2022neural} present a projected neural network method, using differential equations, to tackle a range of sparse optimization problems. This method combines a non-smooth convex loss with L0-norm regularization and proved its global existence, uniqueness, and convergence properties. Some studies employ either L1-norm, L2-norm, or a combination of both for feature selection \cite{ng2004feature}. Indeed, the properties of L0 regularization make it particularly apt for feature selection. Consequently, our method leverages the principles of L0 regularization to design a gating module. This module is utilized to select input variables during the SR training process, as opposed to network features, which were the focus in the aforementioned studies.

\subsubsection{Entropy regularization in reinforcement learning} Enhancing the exploration capabilities during the RL training process can significantly boost the performance of learned policy \cite{ecoffet2019go}. Mnih et al. \cite{mnih2016asynchronous} found that adding the policy entropy to the objective function improved exploration by discouraging premature convergence to suboptimal deterministic policies. Haarnoja et al. \cite{haarnoja2018soft} presented a maximum entropy RL algorithm, which added the policy entropy to the expectation of reward to addresses the challenges of high sample complexity and hyperparameter sensitivity common in RL algorithms. The incorporation of policy entropy as a regularization within the objective function has progressively been established as a standard operational procedure for the RL algorithms \cite{mysore2022multicritic, vinyals2019grandmaster}. To address the early commitment phenomenon and from initialization bias in sequence search of SR, Landajuela et al. \cite{landajuela2021improving} introduced a hierarchical entropy regularizer to enhance the entropy of early actions in sequences. To enhance the exploration of entire sequences rather than individual actions, we introduce the MPE approach, which increases the diversity of generated expressions by amplifying the uncertainty in the generation of complete sequences.

\begin{figure*}[t]
	\centering
	\small
	\includegraphics[scale=0.102]{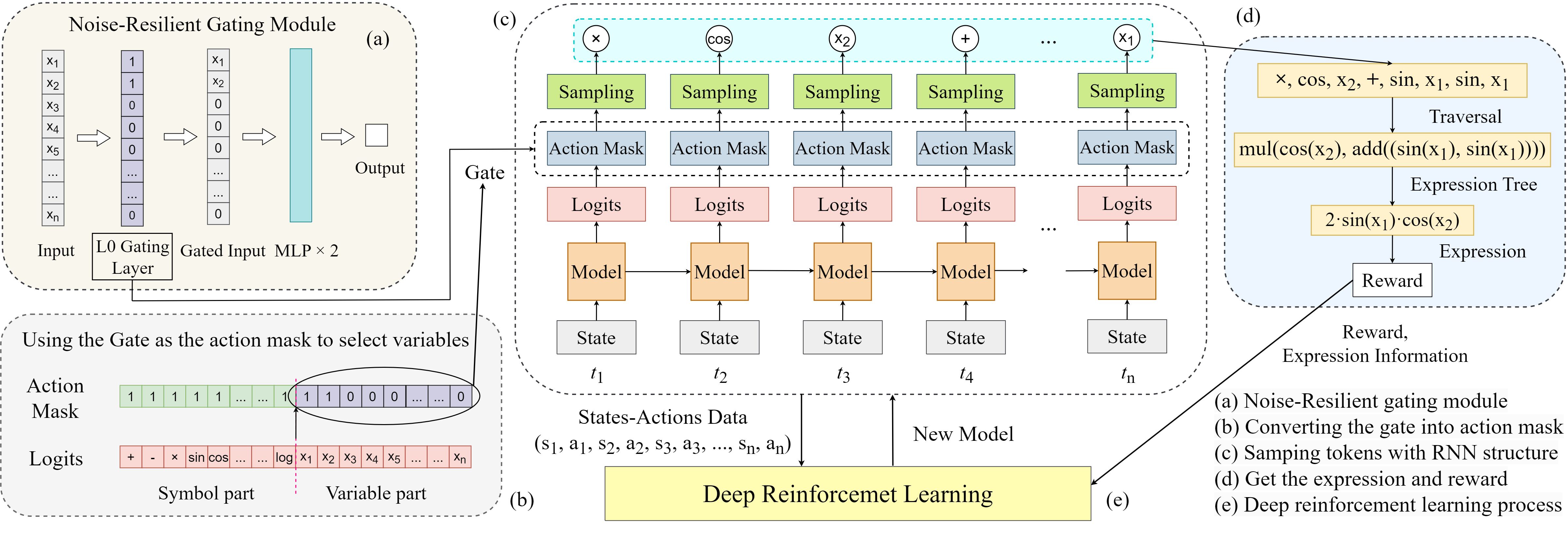}
	\caption{Overview of the NRSR training process. (a) Prior to the SR and RL training process, the NGM is trained with a sample network structure. (b) The obtained L0 gates are then combined with the original action mask to select the input variables. (c) During the SR process, the RNN model, serving as the policy, generates output logits which are processed with the new action mask. These processed logits are subsequently converted into action probabilities, which are used to sample symbolic tokens. The policy generates actions in a step-by-step manner (from $t_0$ to $t_n$) following a time sequence structure, thereby forming a trajectory. (d) Each trajectory, consisting of sequential tokens, represents a traversal that can form an expression using the expression tree approach. The resulting expressions are used to calculate the fitness reward. The training process concludes once the optimal expression is found. (e) The samples, comprising states, actions, and rewards, are used to train a new RL policy. After each training iteration, the updated model is used to generate new expressions. The procedures in (c), (d), and (e) constitute an iterative process in NRSR training. This iteration continues until the optimal expression is found or the limit of consumed expressions is reached, thereby ensuring a comprehensive exploration of the solution space.}
	\label{fig:1}
\end{figure*}

\section{Method}
\label{Section:3}
In this section, we first introduce the designed NGM and how it can filter the noisy input variables. Subsequently, we present the expression generation process during the training process. We then describe the expression generation policy with RL approach, and finally, we introduce the MPE regularization with the RL training process.

\subsection{Noise-Resilient Gating Module}
\label{Section:3.1}
The NGM is designed to induce sparsity in the input variables, effectively performing variable selection by deactivating noisy input as shown in Fig.\ref{fig:1} (a). Given an original input $ X \in \mathbb{R}^{m \times n}$, a L0 gating layer is utilized to selectively filter the input variables to achieve a noise-resilient function. The gating layer $ G \in \{0, 1\}^{m \times n}$ is composed of binary values, where 0 indicates the suppression of the corresponding input and 1 indicates its retention. The processed input $X^{\prime}$ is computed as the Hadamard product of $X$ and $G$:
\begin{equation}
	\label{eq1}
	X^\prime = X \odot G
\end{equation}
where $\odot$ denotes the element-wise multiplication. The network output $Y^\prime$ in the gating module is then computed as:
\begin{equation}
	\label{eq2}
	Y^\prime = W \cdot X^\prime
\end{equation}
where $W$ representing the weights of the network.

The gating layer can be trained by achieving the objective, that is to minimize the mean squared error (MSE) between the network output $Y^{\prime}$ and the true labels $y$, subject to the L0 regularization constraint on $G$. The optimization problem can be formulated as:
\begin{equation}
	\label{eq3}
	J(W,G) = \mathop{min}\limits_{W,G} \dfrac{1}{m} \sum_{i=1}^{m}(y_i-WX^{\prime}_i)^2 + \lambda \left\lVert G\right\rVert_0
\end{equation}
where $\left\lVert G\right\rVert_0$ denotes the L0 norm of $G$, which counts the number of non-zero parameters in the gating layer, and $\lambda$ is a regularization parameter that controls the trade-off between the MSE and the sparsity of $G$. The goal of the L0 norm is to maximize the number of zeros in $G$, effectively deactivating the corresponding input variables and thus performing variable selection. The optimization of $\left\lVert G\right\rVert_0$ is computationally intractable due to the non-differentiability and huge possible combinatorial states of $G$. Hence, we employ an approximation approach \cite{louizos2018learning} that allows for gradient-based optimization, after that, we can get the trained $G$ as the gate to filter the input variables during the expression generation process.

\subsection{Integration of Gating Layer with Action Mask}
\label{Section:3.2}
Action mask can prevent the selection of invalid or undesirable actions, thus accelerate the RL training process \cite{tang2020implementing,hou2023exploring}. In our scenario, NRSR trains the noise-resilient gating layer as an action mask \cite{tang2020implementing} to filter out noisy input variables, thereby significantly enhancing exploration efficiency. Action mask is realized by modifying the output logits of policy network, where make the probabilities of masked actions to zero. Hence, it doesn't impact the network's parameters or the backpropagation process, thus preserving the model's learning capability. Upon the convergence of the training phase of NGM, the learned gating layer $G$ is integrated with the original action mask used during the sampling stage, as depicted in Fig. \ref{fig:1} (b). The original action mask was designed to constrain the illogical sample of symbols within an expression to expedite the exploration process. For instance, it prevents the selection of operations that are inverses or descendants of their predecessors, such as avoiding the sequence $log(exp(x))$ or $sin(cos(x))$ in traversals due to its meaningless \cite{petersen2021deep}. The improved action mask inherits the capability to filter out noisy variables, thereby significantly reducing the complexity of the search space. This is achieved by applying the trained gate $G$ to the action mask, which can be represented as: 

\begin{equation}
	\label{eq4}
	A_{new} = A_{original} \odot G
\end{equation}
where $A_{original}$ denotes the original action mask and $A_{new}$ represents the new action mask that incorporates the learned gates. The element-wise multiplication ensures that only the permissible operators and selected variables are activated during the sampling process.

The improved action mask not only maintains the original constraints but also introduces an additional selection based on the relevance of the input variables. This dual functionality facilitates a more focused and efficient search by eliminating unnecessary exploratory steps and concentrating on the most promising regions of the search space. 

\subsection{Generating Expressions as Training Samples}
\label{Section:3.3}
Mathematical expressions can be effectively represented by expression trees, a specific type of binary tree where internal nodes correspond to mathematical operators and terminal nodes represent input variables or constants \cite{petersen2021deep}. An expression tree can be sequentially described using pre-order traversal, thereby enabling the representation of an expression through the pre-order traversal of its corresponding expression tree. The tokens in these pre-order traversals are constituted by mathematical operators, input variables, or constants. These tokens can be selected from a pre-established token library, denoted as $L$. This library encompasses a range of commonly used operators and given variables, such as $ \{+; -; \times; \div; sin; cos; log; exp; x_1; x_2; ... \}$. Consequently, the generation of an expression can be achieved by sequentially producing tokens along the pre-order traversal as show in Fig. \ref{fig:1} (d). One traversal can be represented as $\tau$, with $\tau_i$ denoting the $i_{th}$ token of $\tau$. The length of the traversal is symbolized by $|\tau| = T $. Consequently, the process of generating expressions can be viewed as a sequence generation process, which can be optimized using a recurrent neural network (RNN). The RNN serves as a potent tool for generating sequences that fit expressions, as it can encapsulate all previous information in the generating traversal. The process of generating expressions via the RNN constitutes the policy $\pi$, which can be optimized by the RL algorithm.

We sample tokens from $L$ with probabilities derived from the output of the last layer of RNN with parameters $\theta$. The sampled tokens, arranged in sequence as the traversal $(\tau_1,\tau_2,\ldots,\tau_T)$, can form the mathematical expression as shown Fig. \ref{fig:1} (c). Specifically, the $i_{th}$ output of the RNN passes through a Softmax layer to produce the probability distribution for selecting the $i_{th}$ token $\tau_i$, conditioned on the previously selected tokens $\tau_{1:(i-1)}$. This process is noted as $\pi_\theta(\tau_i|\tau_{1:(i-1)})$. The likelihood of sampling the entire expression is simply the product of the likelihoods of its tokens: $\pi_\theta(\tau)=\pi_\theta(\tau_1)\prod_{i=2}^{T}{\pi_\theta(\tau_i|\tau_{1:(i-1)})}$. Hence, after getting the traversal $(\tau_1,\tau_2,\ldots,\tau_T)$, the expression $f$ is generated as a training sample.

Except for generating expressions as samples, this stage also necessitates the estimation of effects corresponding to these generated expressions. These effects will serve as reward feedback for the RL training process. Given the test dataset $(X; y)$ with size $n$ and learned expression $f$, the normalized root-mean-square error (NRMSE) can be used as an indicator of the fitness of the learned expression $f$, as calculated by $\frac{1}{\sigma_y}\sqrt{\frac{1}{n}\sum_{i=1}^{n}(y_i-f(X_i))^2}$, where $\sigma_y$ is the standard deviation of the target values. A smaller NRMSE signifies a better performance of the learned expression. In the ideal case where NRMSE equals zero, it implies that the underlying patterns within the data have been perfectly discovered, and the corresponding expression is correct.

\subsection{Reinforcement Learning with Mixed Path Entropy (MPE) Regularization}
\label{Section:3.4}
The traversal generation process can be modeled as a standard Markov Decision Process (MDP), where the policy can be learned using RL approaches. We assume that the preceding operators in the traversals, along with their properties, can be treated as \textit{observations} during the generation process. The selection of tokens in the traversals constitutes the \textit{actions} of the policy based on the corresponding observations. As described in the previous section, the \textit{reward}, which serves as the policy objective for SR, is defined by minimizing the NRMSE. Consequently, the token sequences in the generated traversals represent \textit{trajectories}, and each generation process is considered as an \textit{episode} in the MDP.

The Proximal Policy Optimization (PPO) \cite{schulman2017proximal} as a popular policy gradient algorithm \cite{schulman2015trust,Song2020VMPO} can be used to achieve policy objectives. The reward obtained by a trajectory $\tau$, denoted as $R(\tau)$, is calculated as 1 / (1 + NRMSE), which is the inverse of the NRMSE (INRMSE). The reward $R(\tau)$ is assigned upon the completion of an episode and is attributed to each action within that episode. Then the empirical $(1 - \eta)$-quantile of $R(\tau)$ in batch data, represented as $R_\eta$, is used to filter the batch data to only include the top $\varepsilon$ fraction of samples. This approach make the policy focus on learning from best-case expressions, which can increase the training efficiency \cite{petersen2021deep}. The term $R_\eta$ can be regarded as the baseline within the advantage function of PPO, thereby serving as a substitute for the value function in PPO.

Entropy regularization is a popular approach in policy gradient methods to prevent premature policy convergence and to encourage exploration. We employ a hierarchical entropy term to increase the randomness of the policy at each individual step \cite{landajuela2021improving}, which is described as:
\begin{equation}
	\label{eq6}
	H\left(\pi_\theta\right)=-\sum_{\tau\in\Gamma}\sum_{t=1}^{T}{\gamma^{t-1}\sum_{\tau_t\in a}{\pi_\theta\left({\tau_t|s}_t\right)\log{\pi_\theta\left(\tau_t|s_t\right)}}}
\end{equation}
where $\gamma < 1$ is an exponential decay factor for the weights. $\tau_t$ is the action sampled by the policy at the state $s_t$, and $a$ is all possible actions. $H\left(\pi_\theta\right)$ can encourage the exploration of policy, especially in the earliest tokens of each trajectory to alleviate the early commitment phenomenon and initialization bias in symbolic spaces.

To further promote exploration across sequences, we introduce path entropy regularization, $H_\mathrm{\tau}\left(\pi_\theta\right)$, which is a measure of the uncertainty of an entire sequence of actions in all sequences taken by the policy. Given that the objective of RL is to maximize the expectation of reward, during the training of RL-based SR, it is possible to prematurely converge to a local-optimum state that closely approximates the target expression, but with tokens that are entirely absent from the target expression. This could lead to continuous exploration around the local-optimum state. In this case, the likelihood of the policy exploring the correct tokens present in the target expression is significantly low. The aim of path entropy regularization is to facilitate the discovery of the perfect expression that necessitates more exploration of completely distinct sequence paths. Hence, for a given trajectory set $\mathrm{\Gamma}$, the path entropy can be defined as:
\begin{equation}
	\label{eq7}
	H_\mathrm{\tau}\left(\pi_\theta\right)=-\sum_{\tau\in\mathrm{\Gamma}}{\pi_\theta\left(\mathrm{\tau}\right)}\log{\pi_\theta\left(\mathrm{\tau}\right)}
\end{equation}
where $\pi_\theta(\mathrm{\tau})$ is the joint probability of the entire action sequence $(\tau_1,\tau_2,\ldots,\tau_T)$ of trajectory $\tau$, which is computed as:
\begin{equation}
	\label{eq8}
	\pi_\theta\left(\mathrm{\tau}\right)=\pi_\theta\left(\tau_1\right)\prod_{t=2}^{T}{\pi_\theta\left(\tau_t\middle|\tau_{t-1},\ldots,\tau_1\right)}
\end{equation}

Finally, the policy objective that combines both single-step entropy term and path entropy term is formulated as:
\begin{equation}
	\label{eq9}
	\mathcal{L}\left(\theta\right)=\mathcal{L}_p\left(\theta\right)+\alpha H_\mathrm{\tau}\left(\pi_\theta\right)+\beta H\left(\pi_\theta\right)
\end{equation}
where $\alpha$ and $\beta$ are hyperparameters that determines the significance of the path entropy term and single-step entropy term, respectively. This combined entropy term aims to balance the immediate exploration benefits of single-step entropy with the long-term diversity encouraged by path entropy, which can be called mixed path entropy (MPE). The pseudocode of NRSR is shown in Appendix.

\section{Experiments}
\label{Section:4}
In this section, we first delineate the experimental configurations. Following this, we present the results and conduct a comprehensive analysis to validate the efficacy of our method, as well as the individual modules encompassed within it.

\subsection{Experimental Configurations}
\label{Section:4.1}

\subsubsection{Benchmark}
\label{Section:4.1.1}
In this study, we employed the Nguyen SR benchmark suite \cite{uy2011semantically} to assess our proposed method. This suite, widely used in SR research, comprises twelve representative expressions. Each benchmark is defined by a ground truth expression, an operator library, and an input variable range, all of which are detailed in Appendix. Datasets are generated using the ground truth and the input range, and are subsequently divided into three segments: one for training the NGM, one for calculating the fitness reward $R(\tau)$ of the expressions generated during the training process, and one for evaluating the best fit expression after each training iteration. The sample sizes for these three subsets are 20,000, 20, and 20, respectively. The operator library, which restricts the operators available for use during training, is denoted by $ \{+; -; \times; \div; sin; cos; log; exp; x_i \}$ in this study, with the $i_{th}$ input variables represented by $x_i$.

\subsubsection{Baselines}
\label{Section:4.1.2}
We compared NRSR against five sota SR baselines, providing a comprehensive comparison across two RL-based methods, a GP-based method, a pre-trained method and a commercial software. The first baseline, \textbf{DSR} \cite{petersen2021deep}, is a RL-based SR framework that employs an RNN with a risk-seeking PG to generate and optimize mathematical expressions, demonstrating superior performance. The second baseline \cite{landajuela2021improving} builds upon DSR by incorporating a hierarchical-entropy regularizer and a soft-length prior, which is noted by \textbf{HESL} in this paper. This enhancement mitigates early commitment and initialization bias, thereby improving exploration and performance. The third, \textbf{GP-Meld} \cite{landajuela2022unified}, is an SR method that combines GP with DSR. It uses GP as an inner optimization loop, augmenting the exploration of the search space while addressing the non-parametric limitations of GP with DSR's neural network. The fourth baseline, \textbf{DGSR} \cite{holt2023deep}, utilizes pre-trained deep generative models to exploit the inherent regularities of equations, enhancing the effectiveness of SR. It has excellent performance in terms of recovering true equations and computational efficiency, particularly dealing with a large number of input variables. The final baseline is \textbf{Eureqa} \cite{white2012software}, a widely-used commercial software based on a GP-based approach \cite{schmidt2009distilling}, serving as the gold standard for SR.

\subsubsection{Training Process}
\label{Section:4.1.3}
In the SR process for each benchmark, the NGM is initially trained to obtain the noise-resilient gating layer, a process that constitutes a regression task. During the RL training phase, the acquired gates $G$ are employed to filter out noisy input variables. In the expression sampling phase, the RL policy sequentially generates tokens to produce a batch of trajectories. These trajectories are subsequently transformed into expressions to compute the fitness reward $R(\tau)$. For each iteration of policy training, the trajectory data, reward $R(\tau)$, and information about the generated expressions are utilized to train new policies. NRSR and the other comparative methods were implemented within a unified SR framework \cite{landajuela2022unified}, with the exception of Eureqa, which was executed using the API interface of the DataRobot platform\footnote[1]{[Online]. Available: \url{https://docs.datarobot.com}}. Detailed specifications of the training settings can be found in Appendix.

\subsection{Results}
\label{Section:4.2}

\subsubsection{SR Performance on high-noise data}
\label{Section:4.2.1}
The performance of NRSR and five comparative baselines is evaluated using three metrics: recovery rate (RR), explored expression number (EEN), and normalized mean-square error (NMSE). RR quantifies the likelihood of identifying perfect expressions across all replicated tests under varying random seeds. EEN represents the average number of expressions examined across all replicated tests. A lower EEN indicates greater efficiency of a method in discovering the correct expression with fewer training resources. EEN is particularly critical for SR tasks, as they are conjectured to be NP-hard. NMSE measures the average fitness discrepancy between the ground-truth expression and the best-found expression across all replicated tests. Table \ref{table:1} presents the average performance of six SR methods across all benchmarks when applied to high-noise data. Our proposed method, NRSR, significantly outperforms the five baseline methods in terms of RR, EEN, and NMSE. The results highlight a substantial degradation in the performance of the baseline methods when five noisy inputs are introduced. This performance decline becomes even more noticeable as the number of noisy inputs escalates to ten. Contrarily, NRSR maintains high performance levels even in the presence of high-noise data, thereby demonstrating its robustness against high-noise interference. Unless otherwise specified, all results reported in this study are the average of 100 replicated tests, each with different random seeds, for each benchmark expression.

\begin{table}[t]
	\centering
	\small
	\begin{tabular}{lccc}
		\hline
		Methods &  RR $\uparrow$  & EEN $\downarrow$ & NMSE $\downarrow$ \\ \hline
		           \multicolumn{4}{c}{(a) with 5-noise data}             \\ \hline
		DSP     &     61.3\%      &       955K       &      0.0352       \\
		HESL    &     63.9\%      &       895K       &      0.0300       \\
		GP-Meld &     51.0\%      &      1.63M       &      0.0491       \\
		DGSR    &     72.5\%      &       712K       &      0.0101       \\
		Eureqa  &     35.0\%      &       NaN        &       0.176       \\
		NRSR    & \textbf{89.1\%} &  \textbf{425K}   & \textbf{7.73e-3}  \\ \hline
		           \multicolumn{4}{c}{(b) with 10-noise data}            \\ \hline
		DSP     &     23.2\%      &      1.63M       &       0.138       \\
		HESL    &     26.4\%      &      1.58M       &       0.123       \\
		GP-Meld &     35.1\%      &      1.80M       &      0.0718       \\
		DGSR    &     68.5\%      &       864K       &      0.0121       \\
		Eureqa  &     34.3\%      &       NaN        &       0.282       \\
		NRSR    & \textbf{89.1\%} &  \textbf{423K}   & \textbf{8.52e-3}  \\ \hline
	\end{tabular}
	\caption{Comparison of average RR, EEN, and NMSE between NRSR and five baseline methods across all benchmarks in high-noise data scenarios.}
	\label{table:1}
\end{table}

\subsubsection{Ablation Studies}
\label{Section:4.2.2}
Ablation studies were conducted to estimate the individual contributions of the critical components within NRSR. As expounded in Section 3, the NGM and the MPE emerged as pivotal constituents. Furthermore, the efficacy of the PPO algorithm was also appraised against the traditional PG approach. To ensure a precise evaluation of each component's influence, the NGM was specifically tested under noisy data conditions, whereas the performance metrics for the remaining modules were obtained using clean data. These ablation tests were carried out utilizing the Nguyen benchmark suite, employing RR, EEN, and NMSE as the metrics, with the results presented in Table \ref{table:2}. The outcomes demonstrate that the NGM significantly enhances SR performance on high-noise data scenarios. The omission of MPE and PPO results in diminished performance across both high-noise and clean data contexts, representing the advantageous role of MPE and PPO in augmenting the RL training process for SR tasks. In the ablation tests, the high-noise data have 10 noisy input variables. 

\begin{table}[t]
	\centering
	\small
	\begin{tabular}{lccc}
		\hline
		Ablations          &  RR $\uparrow$  & EEN $\downarrow$ & NMSE $\downarrow$ \\ \hline
		                \multicolumn{4}{c}{(a) on high-noise data}   \\ \hline
		NRSR               & \textbf{89.1\%} &  \textbf{423K}   & \textbf{8.52e-3}  \\
		No NGM             &     32.9\%      &      1.45M       &       0.093       \\
		No NGM / MPE / PPO &     28.3\%      &      1.62M       &       0.112       \\ \hline
		                   \multicolumn{4}{c}{(b) on clean data}    \\ \hline
		NRSR               & \textbf{89.7\%} &  \textbf{408K}   &      7.66e-3      \\
		No MPE             &     85.3\%      &       431K       & \textbf{5.98e-3}  \\
		No PPO             &     86.7\%      &       462K       &      8.92e-3      \\
		No MPE / PPO       &     84.1\%      &       480K       &      9.60e-3      \\ \hline
	\end{tabular}
	\caption{Comparison of average RR, EEN, and NMSE across all benchmarks for various ablations of NRSR.}
	\label{table:2}
\end{table}

\subsubsection{Analysis of noise-resilient gating module}
\label{Section:4.2.3}
The NGM's performance is crucial for SR in scenarios with high-noise data. Hence, a series of experiments were devised to assess the reliability of the gating mechanism under a variety of parameter configurations. Our evaluation are conducted on 93 expressions, which are detailed in the Supplementary File. We introduced four levels of noisy inputs across a comprehensive set of all benchmarks. The gating layers were then employed to filter the input variables, with the perfect filter rate being the primary metric of interest. The results, presented in Table \ref{table:6}, reveal that the gating layer adeptly eliminates noisy input variables with a 98.92\% accuracy when the number of noise variables does not exceed ten. This mechanism demonstrates robust and stable performance, retaining high accuracy even with the introduction of up to twenty noisy input variables. Further investigation was conducted to compare the efficacy of employing the gating layer extracted from the final training epoch against an averaged gating layer computed across all epochs. The empirical evidence suggests that utilizing the averaged gating layer in conjunction with the Otsu threshold \cite{Otsu1979ATS} significantly surpasses the approach using the gates on the final epoch. This enhancement is likely due to the averaged approach's stability, which mitigates the variability inherent in the results from the final epoch. Moreover, we examined the impact of adjusting the Otsu threshold scale and the L0 gating loss coefficient $\lambda$. The optimal performance was achieved by scaling the Otsu threshold to 1.05 times its original value, facilitating more effective noise filtration. However, further scaling diminishes the benefits, as an excessively high threshold risks discarding true input variables. It is noteworthy that, with the exception of the first test set, the number of noisy input variables was fixed at ten for these evaluations.

\begin{table}[t]
	\centering
	\small
	\begin{tabular}{lc}
		\hline
		Settings                                     & Accuracy \\ \hline
		with 3 noisy input variables                 & 98.92\%  \\ 
		with 5 noisy input variables                 & 98.92\%  \\ 
		with 10 noisy input variables                & 98.92\%  \\ 
		with 20 noisy input variables                & 96.77\%  \\ 
		using the gating layer on the final epoch 	 & 60.21\%  \\ 
		Otsu threshold scale 1.0                     & 95.70\%  \\ 
		Otsu threshold scale 1.05                    & 98.92\%  \\ 
		Otsu threshold scale 1.1                     & 94.62\%  \\ 
		gating loss $\lambda$ 0.1                 & 96.77\%  \\ 
		gating loss $\lambda$ 0.25                & 98.92\%  \\ 
		gating loss $\lambda$ 0.5                 & 96.77\%  \\ \hline
	\end{tabular}
	\caption{Comparison of accuracy results under different training settings for NGM.}
	\label{table:6}
\end{table}

\subsubsection{Analysis of mixed path entropy}
\label{Section:4.2.4}
In the context of SR, the MPE is instrumental in promoting the exploration of diverse complete sequences. To augment our understanding of MPE's efficacy in SR, we introduce a metric named as the effective exploration ratio (EER). EER is calculated as the ratio of the unique expression number (UEN) to the EEN, wherein UEN is the amount of unique expressions generated through successful explorations. If the exploration fails to accurately reconstruct the target expression, the corresponding UEN is recorded as zero. EEN represents the amount of expression generated per task. Hence, a higher EER value indicates a more efficient exploration process. The evaluation of EER across NRSR and four comparative baselines, conducted on the Nguyen-5 benchmark, is presented in Table \ref{table:7}. The results indicate that NRSR has the highest EER, suggesting its superior proficiency in exploration efficiency within the SR framework.

\begin{table}[b]
	\centering
	\setlength{\tabcolsep}{1mm}
	\small
	\begin{tabular}{cccccc}
		\hline
		Matrix &     NRSR     &   DSR   &  HESL  & GP-Meld &  DGSR  \\ \hline
		UEN   &     524.3K     & 718.4K  & 651.7K & 104.7K  & 107.5K \\
		EEN   &     735.2K     & 1132.4K & 988.1K & 1853.7K & 195.6K \\
		EER   & \textbf{0.713} &  0.634  & 0.660  &  0.056  & 0.550  \\ \hline
	\end{tabular}
	\caption{Comparison of UEN, EEN, and EER results between our proposed method and four baselines on the Nguyen-5 benchmark.}
	\label{table:7}
\end{table}

\begin{figure}[t]
	\centering
	\small
	\includegraphics[width=0.23\textwidth]{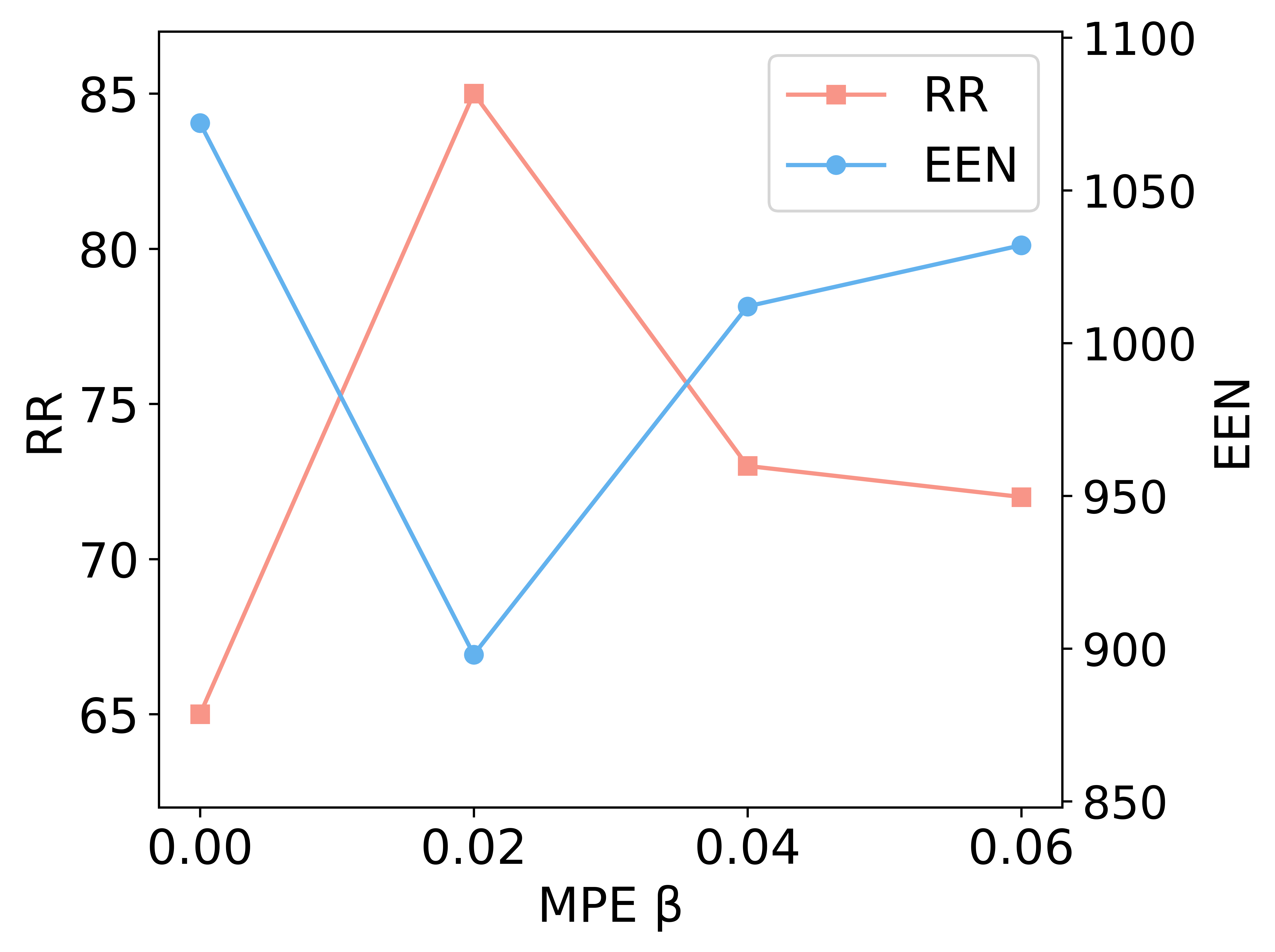}
	\includegraphics[width=0.23\textwidth]{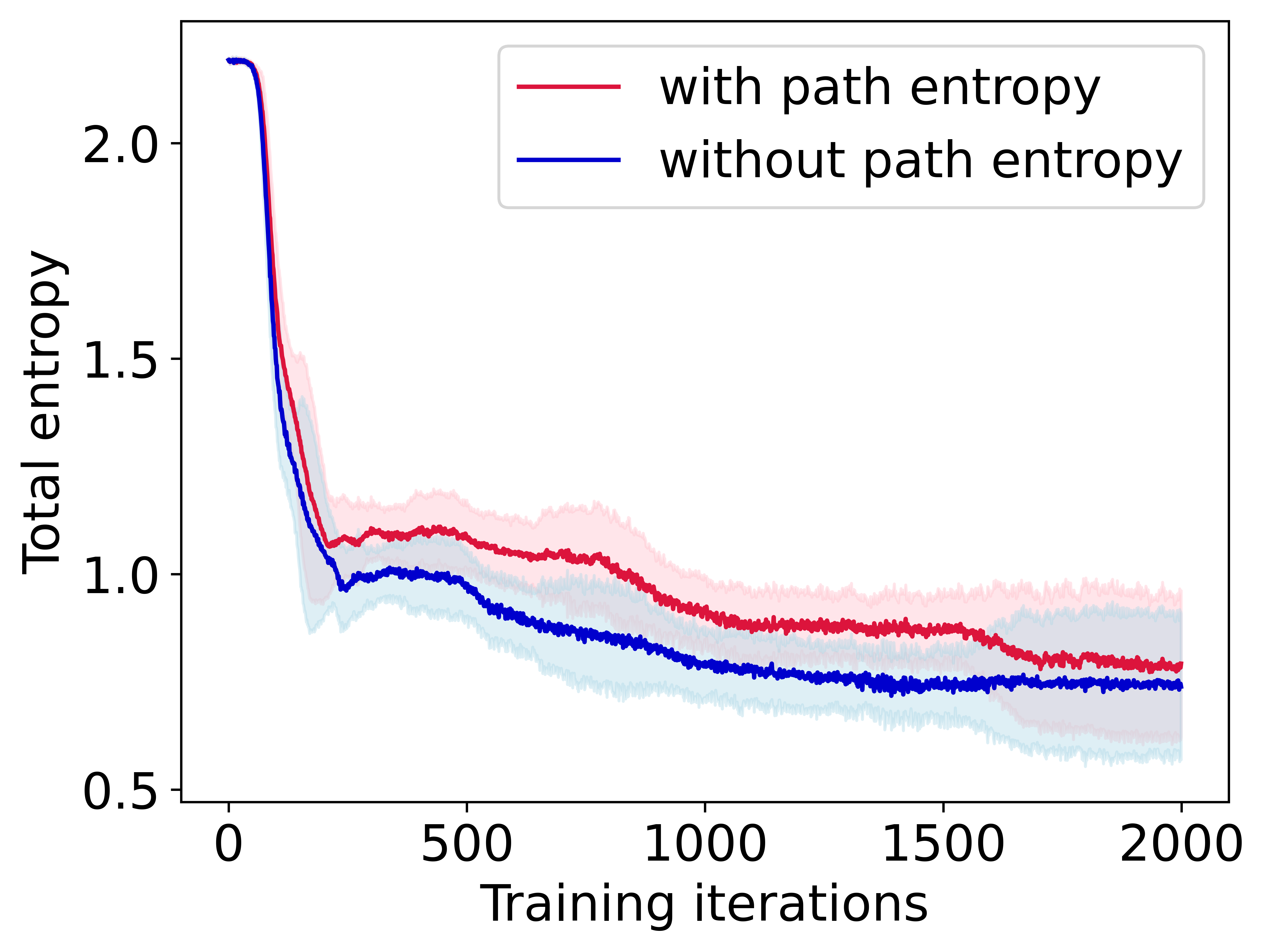}
	\caption{(a) Variation of RR and EEN metrics performance with respect to $\beta$ in MPE. (b) Dynamics of total entropy during the training phase.}
	\label{fig2}
\end{figure}

To analyze the impact of the MPE on SR, we conducted a series of tests adjusting the parameter $\beta$ within the MPE on the Nguyen-7 benchmark. The outcomes are represented in Fig. \ref{fig2}(a). The $\beta$ is incremented from 0 to 0.06, there is a notable inflection in the performance. Specifically, the RR exhibits an initial ascent, followed by a descent. Concurrently, the EEN manifests a converse trend, initially presenting a decline, which then transitions into an ascent. This pattern implies a trade-off inherent in the MPE: it has the potential to improve exploratory behavior, yet an overly large $\beta$ may heighten the computational expenditure of the algorithm. Further insights are provided in Fig. \ref{fig2}(b), which illustrates the dynamics of total entropy during the training phase on the Nguyen-12 benchmark. It demonstrates that the employment of MPE will result in a higher total entropy compared to excluding path entropy throughout the training duration, while simultaneously maintaining superior SR performance. These results were derived from ten independent tests, each conducted with a random seed. 

\subsubsection{SR Performance on clean data}
\label{Section:4.2.5}
To thoroughly assess the performance of our proposed NRSR, we conducted the benchmark test against established baseline methods using a dataset without noises. The results, detailed in Table \ref{table:8}, show the comparative effectiveness of various SR methods on clean data. Notably, NRSR outperforms the five baseline methods on average RR and EEN. This demonstrates NRSR's robustness and its ability to deliver superior performance as an independent SR method on clean data. Ablation studies partly illuminate the sources of this improved performance of our proposed method. Specifically, the adoption of the PPO algorithm and the MPE bonus are identified as significant contributors to the method's success. While Eureqa exhibits superior performance in terms of NMSE, it registers the lowest RR. This discrepancy may stem from the employed algorithms in Eureqa.

\begin{table}[t]
	\centering
	\small
	\begin{tabular}{lccc}
		\hline
		Methods &  RR $\uparrow$  & EEN $\downarrow$ & NMSE $\downarrow$ \\ \hline
		DSP     &     83.2\%      &       540K       &      8.75e-3      \\
		HESL    &     88.3\%      &       441K       &      7.94e-3      \\
		GP-Meld &     80.2\%      &       879K       &      5.75e-3      \\
		DGSR    &     77.7\%      &       507K       &      1.19e-2      \\
		Eureqa  &     67.4\%      &       NaN        & \textbf{1.97e-3}  \\
		NRSR    & \textbf{89.7\%} &  \textbf{408K}   &      7.66e-3      \\ \hline
	\end{tabular}
	\caption{The comparison of average RR, EEN and NMSE for our proposed method and five baselines on all benchmarks with clean data.}
	\label{table:8}
\end{table}

\section{Conclusion}
\label{Section:6}
We present an innovative symbolic regression (SR) method that demonstrates excellent capability to precisely recover expressions from data with high-noise, outperforming existing state-of-the-art (sota) methods on a comprehensive set of benchmark tasks. Our method utilizes a noise-resilient gating module (NGM) to filter out noisy information and employs reinforcement learning approach to develop policies for expression recovery. We also introduce the mixed path entropy (MPE), a novel policy bonus designed to enhance the exploration of expressions during training. Our experimental results demonstrate that our method not only handles high-noise data with superior performance but also achieves sota results on benchmarks with clean data. This shows the robustness of our method in various data quality scenarios. Importantly, the NGM and MPE are designed as modular elements, making them suitable for integration into other SR frameworks, thus expanding their potential utility and contributing to the advancement of SR methodologies.

\section{Acknowledgments}
We would like to express our sincere gratitude to the reviewers for their insightful comments and valuable suggestions. We also thank our colleagues at the Cooperation Product Department, Interactive Entertainment Group, Tencent for their helpful discussions and support throughout this work.

\bibliography{sun}
\newpage
\appendix
\onecolumn

\section{A. Method}
\subsection{A.1. Implementation of L0 Regularization}

The primary objective of $L_0$ regularization is to minimize the number of non-zero elements in the parameter set $\theta$. This can be formalized by introducing an $L_0$ regularization term:
\begin{equation}
	\label{eq1}
	L_0(\theta) = \|\theta\|_0 = \sum_{j=1}^{|\theta|} I[\theta_j \neq 0]
\end{equation}
This term is added to the original loss function:
\begin{equation}
	\label{eq2}
	L(\theta) = L_E(\theta) + \lambda L_0(\theta)
\end{equation}
where $L_E(\theta)$ is the original loss function, and $\lambda$ is a regularization parameter that controls the trade-off between the original loss and the sparsity of the parameters. The inclusion of this term directly penalizes the number of non-zero parameters, thereby encouraging sparsity in the model.

To achieve sparsity in the parameters $\theta$, we introduce a mask random variable $Z = \{Z_1, \ldots, Z_{|\theta|}\}$, where each $Z_j$ follows a Bernoulli distribution with parameter $q_j$. This allows us to rewrite the $L_0$ regularization term as:
\begin{equation}
	\label{eq3}
	L_0(\theta, q) = \mathbb{E}_{Z \sim \text{Bernoulli}(q)} \left[ \sum_{j=1}^{|\theta|} Z_j \right] = \sum_{j=1}^{|\theta|} q_j
\end{equation}
Introducing mask variables allows us to control which parameters are active (non-zero) and which are inactive (zero), facilitating sparsity.

The original loss function $L_E(\theta)$ is redefined to include the expectation over the mask random variables:
\begin{equation}
	\label{eq4}
	L_E(\theta, q) = \mathbb{E}_{Z \sim \text{Bernoulli}(q)} \left[ \frac{1}{N} \left( \sum_{i=1}^N L(\text{NN}(x_i; \theta \odot Z_i), y_i) \right) \right]
\end{equation}
The total loss function then becomes:
\begin{equation}
	\label{eq5}
	L(\theta, q) = L_E(\theta, q) + \lambda L_0(q)
\end{equation}
By incorporating the expectation over the mask variables, we can account for the stochastic nature of the mask in the loss function, ensuring that the model is trained with the consideration of sparsity.

Since Bernoulli sampling is inherently non-differentiable with respect to \( q \), we employ the Gumbel-Softmax trick \cite{jang2017categorical} to reparameterize the sampling process, thereby making it differentiable. This reparameterization involves leveraging the properties of the Gumbel distribution to transform the sampling process into a differentiable one. Specifically, we replace the non-differentiable \( \text{argmax} \) operation in the Gumbel-Max trick \cite{NEURIPS2021_5b658d2a} with a differentiable softmax operation, yielding a continuous approximation of the discrete distribution. This approach allows us to make the sampling process differentiable, which is essential for gradient-based optimization.

During the sampling process, we employed the Binary Concrete distribution to further enhance sparsity, which begins by sampling \( X \) from the Binary Concrete distribution, defined as:
\begin{equation}
	\label{eq6}
	X \sim \text{BinConcrete}(\alpha, \tau) = \sigma \left( \frac{\ln U - \ln (1 - U) + \ln \alpha}{\tau} \right)
\end{equation}
where \( \sigma \) denotes the sigmoid function, \( U \) is a uniform random variable, \( \alpha \) is a parameter, and \( \tau \) is the temperature parameter. Next, we stretch and shift \( X \) using the transformation:
\begin{equation}
	\label{eq7}
	\bar{X} = X(b - a) + a
\end{equation}
where \( a \) and \( b \) are constants that define the range of the transformation. Finally, we apply the hard-sigmoid function to clip \( \bar{X} \) to the \((0,1)\) interval, resulting in the gating random variable \( Z \):
\begin{equation}
	\label{eq8}
	Z = \min(1, \max(0, \bar{X}))
\end{equation}
This distribution effectively promotes sparsity by making the sampled values closer to binary (0 or 1), thereby enhancing the model's ability to perform gradient-based optimization.

To compute the probability that the gating random variable \( Z \) is non-zero using the Binary Concrete distribution, we use the following equation:
\begin{equation}
	\label{eq9}
	P(Z \neq 0) = \sigma \left( \ln \alpha - \tau \ln \left( -\frac{a}{b-a} \right) \right)
\end{equation}

The \( L_0 \) regularization term, which quantifies the sparsity by computing the probability that each parameter is non-zero, is given by:
\begin{equation}
	\label{eq10}
	L_0(\varphi) = \sum_j P(Z_j \neq 0 | \varphi_j) = \sum_j \sigma \left( \ln \alpha_j - \tau_j \ln \left( -\frac{a}{b-a} \right) \right)
\end{equation}
where \( \varphi_j \) represents the parameters associated with the \( j \)-th gating variable.

To incorporate the \( L_0 \) regularization term into the total loss function, we combine it with the original loss function \( L_E(\theta, \varphi) \):
\begin{equation}
	\label{eq11}
	L(\theta, \varphi) = L_E(\theta, \varphi) + \lambda L_0(\varphi)
\end{equation}
By following these steps, \( L_0 \) regularization is effectively implemented, leading to a sparse and regularized neural network model. This approach ensures that the model not only fits the data well but also remains sparse, which can improve generalization and reduce overfitting.

\subsection{A.2. Markov Decision Process (MDP)}
Reinforcement Learning (RL) is typically formulated as a Markov Decision Process (MDP), defined by the tuple $\langle S, A, P, r, \gamma \rangle$. Here, $S$ represents the state space, and $A$ denotes the action space. The transition function $P: S \times A \times S \rightarrow [0, 1]$ captures the environment dynamics, specifying the probability of transitioning to state $s_{t+1} \in S$ from state $s_t \in S$ by taking action $a \in A$. The reward function $r: S \times A \rightarrow \mathbb{R}$ assigns a reward to each state-action pair. A policy $\pi(a|s)$ is the agent’s behavior function, mapping states to actions or providing a probability distribution over actions. The value function $V^\pi(s)$ evaluates the quality of a state by predicting future rewards. The dynamics model of the environment is a function that predicts the next state and reward given the current state and action. An RL agent may consist of one or more of these components. In RL, the goal is to learn an optimal policy $\pi^\ast$ that maximizes the expected discounted sum of rewards. Formally, the optimal policy is defined as:
\[
\pi^\ast = \arg\max_\pi \mathbb{E}_s \left[ V^\pi(s) \right]
\]
where the value function $V^\pi(s)$ is given by:
\[
V^\pi(s) = \mathbb{E}_{\tau \sim \pi, P(s)} \left[ \sum_{t=0}^{\infty} \gamma^t r(s_t, a_t) \right]
\]
Here, $\tau \sim \pi$ with $P(s)$ indicates sampling a trajectory $\tau$ for a horizon $T$ starting from state $s$ using policy $\pi$ in the MDP with transition model $P$, and $s_t \in \tau$ is the $t$-th state in the trajectory $\tau$.

\subsection{A.3. Proximal Policy Optimization (PPO)}
The Proximal Policy Optimization (PPO) \cite{schulman2017proximal} as a popular policy gradient algorithm \cite{schulman2015trust,Song2020VMPO} can be used to achieve policy objectives. The importance ratio of policy in the PPO takes the form: $r_t=\frac{\pi_\theta\left(\tau_t\middle| s_t\right)}{\pi_{\theta^\prime}\left(\tau_t\middle| s_t\right)}$, where $s_t$ and $\tau_t$ represent state and action in trajectory $\tau$ at sequence step $t$, respectively. $\pi_\theta\left(\tau_t\middle| s_t\right)$ and $\pi_{\theta^\prime}\left(\tau_t\middle| s_t\right)$ are the current and old policies, respectively. Hence, the policy objective is described as:
\begin{equation}
	\label{eq5}
	\mathcal{L}_p\left(\theta\right)=-{\hat{E}}_t\left[min\left(r_t{\hat{A}}_t,clip\left(r_t,1-\varepsilon,1+\varepsilon\right){\hat{A}}_t\right)\right]
\end{equation}
where $\hat{E}\left[\ldots\right]$ indicates the expectation over a finite batch of samples, ${\hat{A}}_t$ is the advantage of policy at sequence step $t$, which is calculated by $R(\tau) - R_\eta$. $\varepsilon$ is the clip hyperparameter. 

The $R_\eta$ can be regarded as the baseline within the advantage ${\hat{A}}$, which can substitute the value function in the PPO.

\subsection{A.3. Pseudocode of NRSR}

The pseudocode of NRSR is shown in Algorithm \ref{alg1}, and the source codes can be found in the \url{https://github.com/clivesun01/nrsr}.

\begin{algorithm}[H]
	\caption{Noise-Resilient Symbolic Regression (NRSR)}
	\begin{algorithmic}
		\STATE 
		\STATE \textbf{factors} Gating layer $G$, NGM parameters $W$, NGM objective $J(G,W)$, RNN parameters $\theta$, RL training objective $\mathcal{L}\left(\theta\right)$, Best fitting expression $\tau^*$, reward function $R(\tau)$, batch size $N$
		\STATE \textbf{for} each NGM training iteration \textbf{do}
		\STATE \hspace{0.5cm} Perform gradient step on $G$ and $W$ by minimizing the NGM objective $J(G,W)$
		\STATE \textbf{end for}
		\STATE \textbf{for} each RL training iteration \textbf{do}
		\STATE \hspace{0.5cm} Sample $N$ expressions by $\pi_\theta$ with $G$, and get $(s_t^\tau, a_t^\tau, R(\tau))$ for each expression $\tau$
		\STATE \hspace{0.5cm} Compute the baseline $R_\eta$ and select top $\eta$ fraction of samples
		\STATE \hspace{0.5cm} Perform gradient step on $\theta$ by maximizing the RL objective $\mathcal{L}\left(\theta\right)	$
		
		\STATE \hspace{0.5cm} \textbf{if} $max{R\left(\tau^i\right)}_{i=1}^N == 1$ \textbf{then}
		\STATE \hspace{1cm} $\tau^\ast=argmax\left(R\left(\tau^i\right)\right)$
		\STATE \hspace{1cm} \textbf{return} $\tau^*$
		
		\STATE \hspace{0.5cm} \textbf{if} $max{R\left(\tau^i\right)}_{i=1}^N>R\left(\tau^\ast\right)$ \textbf{then}
		\STATE \hspace{1cm} $\tau^\ast=argmax\left(R\left(\tau^i\right)\right)$
		\STATE \hspace{0.5cm} \textbf{end if}
		\STATE \textbf{end for}
		\STATE \textbf{return} $\tau^*$
	\end{algorithmic}
	\label{alg1}
\end{algorithm}

\section{B. Experiments}

\subsection{B.1. Benchmark}
Table \ref{table:a1} shows the benchmarks employed for the evaluation of our proposed method. These benchmarks encompass 12 representative expressions from the Nguyen SR benchmark suite. Each benchmark is characterized by a ground truth expression and a input variable range. The ground truth and data range can be used to generate the clean dataset. An operator library restricts the available operators for use in the training, with input variables denoted by $x_i$ in this study. The used library for those benchmarks is $ \{+; -; \times; \div; sin; cos; log; exp; x_i \}$. The dataset with noisy input variables can be generated by adding input variables to an additional clean dataset. It is essential to ensure that the noisy input variables remain within the specified data range.

\begin{table}[h]
	\centering
	\small
	\begin{tabular}{lll}
		\hline
		Benchmark & Ground truth expression     & Input variable Range   \\ \hline
		Nguyen-1  & $x_1^3+x_1^2+x_1$                                     & $U(-1; 1)$ \\
		Nguyen-2  & $x_1^4+x_1^3+x_1^2+x_1$                               & $U(-1; 1)$ \\
		Nguyen-3  & $x_1^5+x_1^4+x_1^3+x_1^2+x_1$                         & $U(-1; 1)$ \\
		Nguyen-4  & $x_1^6+x_1^5+x_1^4+x_1^3+x_1^2+x_1$                   & $U(-1; 1)$ \\
		Nguyen-5  & $\sin{\left(x_1^2\right)}\cos{\left(x_1\right)}-1$    & $U(-1; 1)$ \\
		Nguyen-6  & $\sin{\left(x_1\right)}+\sin{\left(x_1+x_1^2\right)}$ & $U(-1; 1)$ \\
		Nguyen-7  & $\log{\left(x_1+1\right)}+\sin{\left(x_1^2+1\right)}$ & $U(-1; 1)$  \\
		Nguyen-8  & $\sqrt{x_1}$                                          & $U(0; 4)$  \\
		Nguyen-9  & $\sin{\left(x_1\right)}+\sin{\left(x_2^2\right)}$     & $U(0; 1)$  \\
		Nguyen-10 & $2\sin{\left(x_1\right)}\cos{\left(x_2\right)}$       & $U(0; 1)$  \\
		Nguyen-11 & ${x_1}^{x_2}$                                         & $U(0; 1)$  \\
		Nguyen-12 & $x_1^4-x_1^3+0.5x_2^2-x_2$                            & $U(0; 1)$  \\ \hline
	\end{tabular}
\caption{The Nguyen benchmark suite is employed to evaluate the effectiveness of SR approaches.}
\label{table:a1}
\end{table}

\subsection{B.2. Training Settings}

In each benchmark, a noise-resilient gating layer is trained to filter out noisy input variables. The gating layers, derived from all epochs on the validation set, are averaged to yield a vector of continuous values. However, only a binary vector can function as an action mask in the subsequent RL phase. Therefore, the Otsu's method \cite{Otsu1979ATS} is utilized to automatically determine a threshold for binarizing the averaged gating layer, then obtaining the discrete noise-resilient gates $G$. The network architecture of NGM encompasses a gating layer, two dense layers, and an output layer, with batch normalization and ReLU activation applied to the dense layers. The hyperparameters for the NGM's training are specified in Table \ref{table:a2}. If the gates consist solely of zeros, it indicates an incapacity to ascertain whether the input variables are merely noisy or entirely non-redundant. In such cases, no input variables are filtered out. It is commonplace for a clean input to yield a gating layer replete with zeros, as none of the variable values exceed the threshold set by Otsu's method. 
To evaluate the efficacy for training the gating layer, an extensive test was conducted, applying the training process to 93 expressions with different counts of introduced noise variables. The results detailing the training performance of the gating layer with respect to these 93 expressions can be found in Supplement File.

\begin{table}[h]
\centering
\small
\begin{tabular}{ll}
	\hline
	Hyperparameter           & Value       \\ \hline
	Training samples         & 20,000      \\
	Batch size               & 256         \\
	Hidden size              & 128         \\
	Train dataset ratio      & 0.8         \\
	Optimizer                & Adam        \\
	Adam $\beta_1, \beta_2 $ & 0.99, 0.999 \\
	Learning rate            & 0.001       \\
	Epoch number             & 20          \\
	L2 regulation weight     & 1e-5        \\
	L0 gating loss $\lambda$ & 0.25        \\
	Otsu threshold scale     & 1.05        \\ \hline
\end{tabular}
\caption{The hyperparameters for training the NGM.}
\label{table:a2}
\end{table}

In the RL training process, the first step involves using the trained L0 gates $G$ to filter the input variables. During each training iteration, the RL policy, implemented as the RNN module, generates a batch of trajectories by sequentially producing tokens. The RNN component, comprising a single-layer LSTM and a dense layer, is responsible for computing the probability distribution of the possible tokens. The generated trajectories are then converted into expressions to calculate the fitness reward $R(\tau)$. Following this, a sample filter process is applied to the batch of samples to retain only the top-$\epsilon$ performing samples \cite{petersen2021deep}. These selected samples are used to further train the PPO policy. The baseline for the policy objective is set as the empirical (1 - $\epsilon$) quantile of the rewards from the batched samples. The RL training continues until an optimal expression is found or the maximum number of expressions set for exploration during the training is reached.

NRSR and other comparative methods were implemented within a single, unified SR framework \cite{landajuela2022unified}, with the notable exception of Eureqa, which is executed based on the API interface of DataRobot platform\footnote[1]{[Online]. Available: https://docs.datarobot.com}. Table \ref{table:a3} presents the training hyperparameters for both the proposed NRSR and the baseline methods. The selection of hyperparameters for the NGM and the MPE component was conducted through a limited grid search, with the optimal configurations being determined based on preliminary trials. The hyperparameters for the baseline methods adhere to the optimal settings as reported in their respective papers. As a commercial software, Eureqa does not offer user-configurable hyperparameters. INRMSE is used as the reward function for all methods in training process. Eureqa employs the root-mean-square error (RMSE) as its fitness measure, and its other implemented details can refer to the DataRobot. For our experiments, Eureqa was run on datasets comprising $5\times10^4$ instances for training each benchmark to completion. It is important to note that the term ``batch size'' in this context refers to the number of trajectories per training iteration, rather than the number of step-level samples.
All experiments were conducted on a Nvidia A10 GPU and 32-core CPU with Python 3.8.

\begin{table}[h]
\centering

\small
\begin{tabular}{llllll}
	\hline
	Hyperparameter                 & NRSR   & DSR      & HESL & GPMeld  & DGSR   \\ 
	\hline
	Max used samples           & $2\times10^6$ & $2\times10^6$ & $2\times10^6$  & $2\times10^6$ & $2\times10^6$      \\
	Batch size                     & 1000     & 1000     & 1000           & 500      & 300    \\
	Policy Algorithm               & PPO      & PG       & PG             & PG       & PG     \\
	$\epsilon$-risk factor         & 0.05     & 0.05     & 0.05           & 0.02     & /    \\
	LSTM hidden size               & 32       & 32       & 32             & 32       & 32      \\
	Train dataset                  & 20       & 20       & 20             & 20       & 100K \\
	Test dataset                   & 20       & 20       & 20             & 20       & 20    \\
	Optimizer                      & Adam     & Adam     & Adam           & Adam     & Adam   \\
	Learning rate                  & 5e-5     & 5e-5     & 5e-5           & 0.0025   & 0.001   \\
	Entropy coef. $\alpha$         & 0.05     & 0.005    & 0.03           & 0.03     & 0.03   \\
	Entropy coef. $\beta$          & 0.02     & /        & /              & /        & /      \\
	Entropy coef. $\gamma$         & 0.7      & /        & 0.7            & 0.7      & 0.7    \\
	Generic generations            & /        & /        & /              & 25       & /    \\
	Mutation prob.                 & /        & /        & /              & 0.5      & /    \\
	Crossover prob.                & /        & /        & /              & 0.5      & /    \\
	Tournament size                & /        & /        & /              & 5        & /      \\
	\hline
\end{tabular}
\caption{The training hyperparameters of SR part.}
\label{table:a3}
\end{table}

\section{C. Results}

\subsection{C.1. The SR Performance cross all benchmarks on high-noise data}

Table \ref{table:a4} displays the detailed training outcomes of those comparison SR methods when applied to high-noise data. NRSR markedly surpasses these five baselines in RR, EEN, and NMSE across all benchmark expressions, with the sole exception being the Nguyen-12 benchmark, where no method successfully identified the correct expression. NRSR sustains high performance levels with the high-noise data, and the effect of varying the number of noisy inputs on the outcomes is statistically insignificant ($p > 0.05$). This results were obtained by averaging 100 replicated tests with different random seeds for each benchmark expression. The unit of measurement for RR is expressed as a percentage.

\begin{table*}[h]
\small
\setlength{\tabcolsep}{1mm}
\centering
\resizebox{\textwidth}{!}{
	\begin{tabular}{ccccccc}
		\hline
		\multirow{2}{*}{\makecell{Benchmark \\ (5 noises)}} & \multicolumn{1}{c}{NRSR} & \multicolumn{1}{c}{DSP} & \multicolumn{1}{c}{HESL} & \multicolumn{1}{c}{GP-Meld} & \multicolumn{1}{c}{DGSR} & \multicolumn{1}{c}{Eureqa} \\ \cline{2-7}
		& RR / EEN / NMSE  & RR / EEN / NMSE  & RR / EEN / NMSE & RR / EEN / NMSE  & RR / EEN / NMSE  & RR / EEN / NMSE  \\ \hline
		Nguyen-1   & 100/ 45K/ 0  & 89/ 377K/ 1.81e-3 & 99/ 152K/ 1.26e-5 & 100/ 1.14M/ 0 & 100/ 21K/ 0  & 97/ NaN/ 2.56e-5 \\
		Nguyen-2   & 100/ 108K/ 0 & 98/ 241K/ 3.16e-4 & 100/ 184K/ 0  & 40/ 1.84M/ 0.0169 & 100/ 78K/ 0  &  0/ NaN/ 0.0846 \\
		Nguyen-3   & 100/ 144K/ 0 & 98/ 250K/ 6.00e-4 & 100/ 206K/ 0 & 79/ 1.71M/ 7.54e-5 & 100/ 131K/ 0  & 40/ NaN/ 1.14e-3 \\ 
		Nguyen-4   & 100/ 159K/ 0 & 99/ 259K/ 4.21e-5  & 100/ 239K/ 0 & 41/ 1.93M/ 6.82e-3 & 99/ 514K/ 1.60e-7 & 19/ NaN/ 0.182 \\
		Nguyen-5   & 97/ 775K/ 2.65e-5 & 31/ 1.71M/ 0.0171 & 25/  1.74M/ 0.0560  & 1/ 2.00M/ 0.0750 & 100/ 624K/ 0   & 47/ NaN/ 1.32e-3 \\
		Nguyen-6   & 100/ 76K/ 0 & 97/ 216K/ 2.77e-3  & 97/ 197K/ 1.04e-3 & 97/ 1.02M/ 1.35e-4 & 100/ 113K/ 0 & 0/ NaN/ 1.68 \\
		Nguyen-7   & 80/ 899K/ 2.42e-5 & 3/ 2.00M/ 1.84e-3 & 7/ 1.91M/ 9.04e-4 & 2/ 1.99M/ 1.98e-3 & 10/ 1.84M/ 3.46e-5 & 0/ NaN/ 0.0902 \\
		Nguyen-8   & 100/ 143K/ 0 & 8/ 1.97M/ 0.0493 & 13/ 1.84M/ 0.061 & 18/ 1.91M/ 0.0439 & 27/ 1.53M/ 4.65e-3 & 1/ NaN/ 0.0274 \\
		Nguyen-9   & 100/ 122K/ 0 & 100/ 196K/ 0 & 100/ 174K/ 0 & 100/ 1.20M/ 0 & 100/ 25K/ 0 & 100/ NaN/ 0 \\ 
		Nguyen-10  & 99/ 386K/ 5.17e-5  & 98/ 530K/ 2.96e-4 & 99/ 421K/ 1.15e-4 & 56/ 1.68M/ 9.43e-3 &  100/ 140K/ 0  & 16/ NaN/ 1.36e-3 \\
		Nguyen-11  & 93/ 239K/ 3.17e-4 & 15/ 1.71M/ 0.0859 & 27/ 1.68M/ 0.0220 & 78/ 1.14M/ 0.124 &  35/ 1.52M/ 0.0623 & 100/ NaN/ 0 \\
		Nguyen-12  & 0/ 2.00M/ 0.0923 & 0/ 2.00M/ 0.263 & 0/ 2.00M/ 0.218 & 0/ 2.00M/ 0.311 & 0/ 2.00M/ 0.0542  & 0/ NaN/ 4.56e-2 \\ \hline
		Average    & \textbf{89.1}/ \textbf{425K}/ \textbf{7.73e-3} & 61.3/ 955K/ 0.0352 & 63.9/ 895K/ 0.0300 & 51.0/ 1.63M/  0.0491 & 72.5/ 712K/  0.0101  &  35.0/ NaN/ 0.176 \\
		\hline
		\multirow{2}{*}{\makecell{Benchmark \\ (10 noises)}} & \multicolumn{1}{c}{NRSR} & \multicolumn{1}{c}{DSP} & \multicolumn{1}{c}{HESL} & \multicolumn{1}{c}{GP-Meld} & \multicolumn{1}{c}{DGSR} & \multicolumn{1}{c}{Eureqa} \\ \cline{2-7}
		& RR/ EEN/ NMSE  & RR/ EEN/ NMSE  & RR/ EEN/ NMSE & RR/ EEN/ NMSE  & RR/ EEN/ NMSE  & RR/ EEN/ NMSE  \\ \hline
		Nguyen-1   & 100/ 41K/ 0  & 32/ 1.42M/ 0.0203 & 38/ 1.30M/ 0.0199 & 43/ 1.75M/ 5.35e-3 & 100/ 26K/ 0
		& 96/ NaN/ 2.44e-5 \\
		Nguyen-2   & 100/ 116K/ 0 & 35/ 1.38M/ 0.0607 & 37/ 1.34M/ 0.0611 & 88/ 1.70M/ 8.17e-4 & 100/ 67K/ 0 & 0/ NaN/ 0.0969 \\
		Nguyen-3   & 100/ 139K/ 0 & 46/ 1.19M/ 0.0739 & 49/ 1.14M/ 0.0701 & 2/ 2.00M/ 0.0688 & 100/ 118K/ 0 & 39/ NaN/ 3.35e-4 \\ 
		Nguyen-4   & 100/ 159K/ 0 & 27/ 1.53M/ 0.179 & 50/ 1.13M/ 0.178 & 4/ 2.00M/ 0.0136  & 95/ 996K/ 2.44e-3 & 22/ NaN/ 0.765 \\
		Nguyen-5   & 96/ 713K/ 2.82e-5 & 1/ 2.00M/ 0.202 & 2/ 1.99M/ 0.157 & 0/ 2.00M/ 0.0920 & 87/ 1.24M/ 0.0106 & 38/ NaN/ 1.58e-3 \\
		Nguyen-6   & 100/ 77K/ 0 & 30/ 1.45M/ 0.110 & 32/ 1.41M/ 0.107 & 92/ 1.42M/ 2.03e-4 & 100/ 151K/ 0 & 0/ NaN/ 2.20 \\
		Nguyen-7   & 81/ 931K/ 2.08e-5 & 4/ 1.96M/ 0.0771 & 3/ 1.97M/ 0.0877 & 0/ 2.00M/ 8.17e-3 & 6/ 1.95M/ 1.46e-5 & 0/ NaN/ 0.213 \\
		Nguyen-8   & 100/ 166K/ 0 & 6/ 1.98M/ 0.212 & 0/ 2M/ 0.230 & 6/ 1.99M/ 0.122  & 12/ 1.83M/ 6.31e-3 & 1/ NaN/ 0.0357 \\
		Nguyen-9   & 100/ 116K/ 0 & 22/ 1.65M/ 0.0969 & 17/ 1.73M/ 0.106 & 100/ 1.38M/ 0  & 100/ 30K/ 0 & 100/ NaN/ 0 \\ 
		Nguyen-10  & 99/ 347K/ 6.01e-5 & 24/ 1.66M/ 0.0561 & 25/ 1.62M/ 0.0539 &  47/ 1.75M/ 0.0109 & 100/ 211K/ 0 & 15/ NaN/ 2.39e-3 \\
		Nguyen-11  & 93/ 273K/ 1.84e-4 & 51/ 1.35M/ 0.167 & 64/ 1.38M/ 0.114 & 39/ 1.65M/ 9.99e-3 & 22/ 1.78M/ 0.0667   & 100/ NaN/ 0 \\
		Nguyen-12  & 0/ 2.00M/ 0.102 & 0/ 2.00M/ 0.402 & 0/ 2.00M/ 0.297 & 0/ 2.00M/ 0.530 & 0/ 2.00M/ 0.0589 & 0/ NaN/ 0.0665 \\ \hline
		Average    & \textbf{89.1}/ \textbf{423K}/ \textbf{8.52e-3} & 23.2/ 1.63M/ 0.138 & 26.4/ 1.58M/ 0.123 & 35.1/ 1.80M/ 0.0718 & 68.5/ 864K/ 0.0121 & 34.3/ NaN/ 0.282 \\
		\hline
	\end{tabular}
}
\caption{The comparison of RR, EEN and NMSE for NRSR and five baselines on all benchmarks with high-noise data.}
\label{table:a4}
\end{table*}

\subsection{C.2. The SR Performance cross all benchmarks on clean data}

Table \ref{table:a8} shows the detailed comparative results of various SR methods on clean data. Our proposed method outperforms the five baseline methods at most benchmarks on all three matrix. This demonstrates NRSR's robustness and its ability to deliver superior performance as an independent SR method on clean data. The results from the DGSR method suggest a propensity for overfitting, as evidenced by its exceptional performance on benchmarks where it excels, and contrasted with significant underperformance on more challenging benchmarks. This issue may be attributed to the reliance on pre-trained models within the DGSR. Those results are obtained by averaging 100 replicated experiments, each with a different random seed. 

\begin{table*}[h]
\centering
\setlength{\tabcolsep}{1mm}
\small
\resizebox{\textwidth}{!}{
	\begin{tabular}{ccccccc}
		\hline
		\multirow{2}{*}{\makecell{Benchmark}} & \multicolumn{1}{c}{NRSR} & \multicolumn{1}{c}{DSP} & \multicolumn{1}{c}{HESL} & \multicolumn{1}{c}{GP-Meld} & \multicolumn{1}{c}{DGSR} & \multicolumn{1}{c}{Eureqa} \\ \cline{2-7}
		& RR/ EEN/ NMSE  & RR/ EEN/ NMSE& RR/ EEN/ NMSE & RR/ EEN/ NMSE& RR/ EEN/ NMSE & RR/ EEN/ NMSE\\ \hline
		Nguyen-1 & 100/ 43K/ 0 & 100/ 46K/ 0 & 100/ 41K/ 0 &  100/ 74K/ 0  &  100/ 10K/ 0 &  100/ NaN/ 0  \\
		Nguyen-2  & 100/ 111K/ 0 & 100/ 137K/ 0 & 100/ 105K/ 0  &100/ 594K/ 0 &  100/ 10K/ 0  &  100/ NaN/ 0  \\
		Nguyen-3 & 100/ 140K/ 0& 0100/ 159K/ 0 & 100/ 139K/  &  100/ 991K/ 0  &  100/ 16K/ 0  &  80/ NaN/ 5.96e-5\\
		Nguyen-4  & 100/ 155K/ 0& 100/ 160K/ 0 & 100/ 159K/ 0  &  100/ 1.31M/ 0 &  100/ 55K/ 0& 59/ NaN/ 9.36e-5  \\
		Nguyen-5  & 97/ 735K/ 2.57e-5 & 93/ 1.13M/ 3.77e-5 & 96/ 988K/ 2.66e-5&  14/ 1.85M/ 3.66e-3 &  100/ 196K/ 0&  56/ NaN/ 1.61e-5\\
		Nguyen-6  & 100/ 73K/ 0 & 100/ 93K/ 0 & 100/ 77K/ 0 &100/ 195K/ 0 &  100/ 11K/ 0  &100/ NaN/ 0 \\
		Nguyen-7  & 85/ 898K/ 2.47e-5  & 53/ 1.13M/ 7.06e-6 & 73/ 1.01M/ 2.86e-6& 53/ 1.72M/ 1.02e-4 &  3/ 1.98M/ 7.65e-6&67/ NaN/ 3.09e-6\\
		Nguyen-8  & 100/ 119K/ 0& 100/ 205K/ 0 & 100/ 117K/ 0  &  100/ 291K/ 0	& 89/ 393K/ 3.27e-3&1/ NaN/ 0.0031  \\
		Nguyen-9  & 100/ 119K/ 0& 100/ 139K/ 0 & 100/ 116K/ 0  &100/ 442K/ 0 &  100/ 13K/ 0&100/ NaN/ 0 \\
		Nguyen-10 & 100/ 288K/ 0& 66/ 942K/ 1.24e-3 & 100/ 271K/ 0&95/ 933K/ 1.28e-3 &  100/ 94K/ 0  &46/ NaN/ 6.88e-4  \\
		Nguyen-11 & 94/ 213K/ 6.76e-3  & 86/ 334K/ 7.40e-4 & 91/ 262K/ 6.72e-3&100/ 139K/ 0 &40/ 131K/ 5.92e-2  &  100/ NaN/ 0  \\
		Nguyen-12 & 0/ 2.00M/ 0.0852& 0/ 2.00M/ 0.103 & 0/ 2.00M/ 0.0885  &  0/ 2.00M/ 0.064 &0/ 2.00M/ 0.064&0/ NaN/ 9.27e-3  \\ \hline
		Average& \textbf{89.7}/ \textbf{408K}/ 7.66e-3 & 83.2/ 540K/ 8.75e-3 & 88.3/ 441K/ 7.94e-3  &  80.2/ 879K/ 5.75e-3 &  77.7/ 507K/ 1.19e-2& 67.4/ NaN/ \textbf{1.97e-3}  \\ \hline
	\end{tabular}
}
\caption{The comparison of RR, EEN and NMSE for our proposed method and five baselines on Nguyen benchmarks with clean data.}
\label{table:a8}
\end{table*}

\subsection{C.3. Result Discussion}
\label{Section:5}
Results presented in Tables \ref{table:a4} and \ref{table:a8} reveals that SR performance tends to decline when dealing with data that has a high-noise, as opposed to clearer, less noisy data. This trend suggests that the presence of noisy input variables adversely affects the quality of SR. Furthermore, methods based on RL appear to yield better SR results compared to those based on GP, although GP-based methods has better performance in certain specific benchmarks. However, it is important to note that GP-based methods are less computationally efficient, typically requiring an order of magnitude more training time than RL-based methods, which has a considerable impact on the speed of SR. Besides, the results shows that the DGSR method exhibits a degree of robustness to noisy inputs. This resilience is attributed to the DGSR's strategy of training distinct pre-trained models tailored to varying input numbers. For instance, in the presence of five additional noise inputs, the models pre-trained for six or seven inputs are employed for evaluation in this study. Conversely, for clean inputs, the models pre-trained for one or two inputs are utilized. However, this approach need the training of separate pre-trained models for different input scenarios, which significantly escalates resource consumption.

Our observations highlight that reconstructing complex expressions remains a challenge, as demonstrated by the uniform inability of all tested methods to successfully recover the Nguyen-12 benchmark. This issue is likely rooted in the length of the sequences and the inherent characteristics of the expressions involved. To illustrate, expressing $x_1^4-x_1^3+0.5x_2^2-x_2$ would necessitate a complex sequence of tokens such as [sub, add, sub, mul, mul, $x_1$, $x_1$, mul, $x_1$, $x_1$, mul, mul, $x_1$, $x_1$, $x_1$, mul, div, $x_2$, add, $x_2$, $x_2$, mul, $x_2$, $x_2$, $x_2$]. Furthermore, the RL-based SR method encounters the challenge of sparse reward issue \cite{hare2019dealing}, meaning that no feedback is provided until the terminate of the entire episode. Hence, the SR method necessitates further enhancement of the model's exploration capabilities. We posit that a distributed RL framework holds significant potential to address this issue. The distributed training paradigm enables the concurrent utilization of thousands of workers to explore expressions, with their insights collectively informing a central learner \cite{espeholt2018impala}. By adopting a distributed training architecture and initializing with random parameter values, we can substantially broaden the exploration space. This approach can markedly increase the likelihood of accurately recovering complex expressions, presenting a promising direction for SR studies.

In our study, each benchmark employed distinct RL policies. For example, two separate models were trained to deduce the expressions for the Nguyen-1 and Nguyen-2 benchmarks. However, the underlying patterns of these two expressions are similar. Besides, expressions often share commonalities that could be learned by a single model. Thus, leveraging a single model to approximate multiple expressions presents an exciting avenue for our future research. One approach could involve training a foundational model and subsequently adapting it to various expressions using techniques such as LoRa \cite{hu2022lora}.

\end{document}